\documentclass[10pt,twocolumn,letterpaper]{article}

\usepackage{cvpr}
\usepackage{times}
\usepackage{epsfig}
\usepackage{graphicx}
\usepackage{amsmath}
\usepackage{amssymb}
\usepackage{booktabs}
\usepackage{multirow}
\usepackage{colortbl}
\usepackage[bottom]{footmisc}
\definecolor{mygray}{gray}{.9}
\newcommand*{\affaddr}[1]{#1} 
\newcommand*{\affmark}[1][*]{\textsuperscript{#1}}
\newcommand*{\email}[1]{\texttt{#1}}

\usepackage[pagebackref=true,breaklinks=true,pdftex,letterpaper=true,colorlinks,citecolor=blue,bookmarks=false]{hyperref}

\cvprfinalcopy 

\def\cvprPaperID{800} 

\ifcvprfinal\fi
\begin{document}    

\title{Search to Distill: Pearls are Everywhere but not the Eyes}

\author{%
\vspace{.1cm}
Yu Liu\thanks{This work is performed in Yu's internship at Google Inc.}\ \affmark[1,3] \ \ \ \ Xuhui Jia\affmark[1]\ \ \ \ Mingxing Tan\affmark[2]\ \ \ \ Raviteja Vemulapalli\affmark[1]\ \ \ \ Yukun Zhu\affmark[1]\\
\vspace{.2cm}
Bradley Green\affmark[1]\ \ \ \ Xiaogang Wang\affmark[3]\\
\vspace{.1cm}
\affaddr{\affmark[1]Google AI} \ \ \ \affaddr{\affmark[2]Google Brain} \ \ \ \\
\affaddr{\affmark[3]Multimedia Laboratory, The Chinese University of Hong Kong}\\
\email{liuyuisanai@gmail.com} \\
\email{\{xhjia, tanmingxing, ravitejavemu\}@google.com}
}

\maketitle
\thispagestyle{empty}


\begin{abstract}
Standard Knowledge Distillation (KD) approaches distill the knowledge of a cumbersome teacher model into the parameters of a student model with a pre-defined architecture. However, the knowledge of a neural network, which is represented by the network's output distribution conditioned on its input, depends not only on its parameters but also on its architecture.\ Hence, a more generalized approach for KD is to distill the teacher's knowledge into both the parameters and architecture of the student. To achieve this, we present a new \textit{Architecture-aware Knowledge Distillation (AKD)} approach that finds student models (pearls for the teacher) that are best for distilling the given teacher model. In particular, we leverage Neural Architecture Search (NAS), equipped with our KD-guided reward, to search for the best student architectures for a given teacher. Experimental results show our proposed AKD consistently outperforms the conventional NAS plus KD approach, and achieves state-of-the-art results on the ImageNet classification task under various latency settings. Furthermore, the best AKD student architecture for the ImageNet classification task also transfers well to other tasks such as million level face recognition and ensemble learning.

\end{abstract}


\section{Introduction}

Over the past few years, Deep Neural Networks (DNNs) have achieved unprecedented success on various tasks, such as image understanding \cite{NIPS2012_4824,liu2017quality}, image generation~\cite{liu2018exploring}, machine translation \cite{wu2016googles} and speech recognition \cite{hinton2012deep}. In the deep learning era, one of the main driving forces for performance improvement is designing better neural architectures \cite{he2015deep,szegedy2016inceptionv4}.

\begin{figure}
\centering
\includegraphics[width=.9\linewidth]{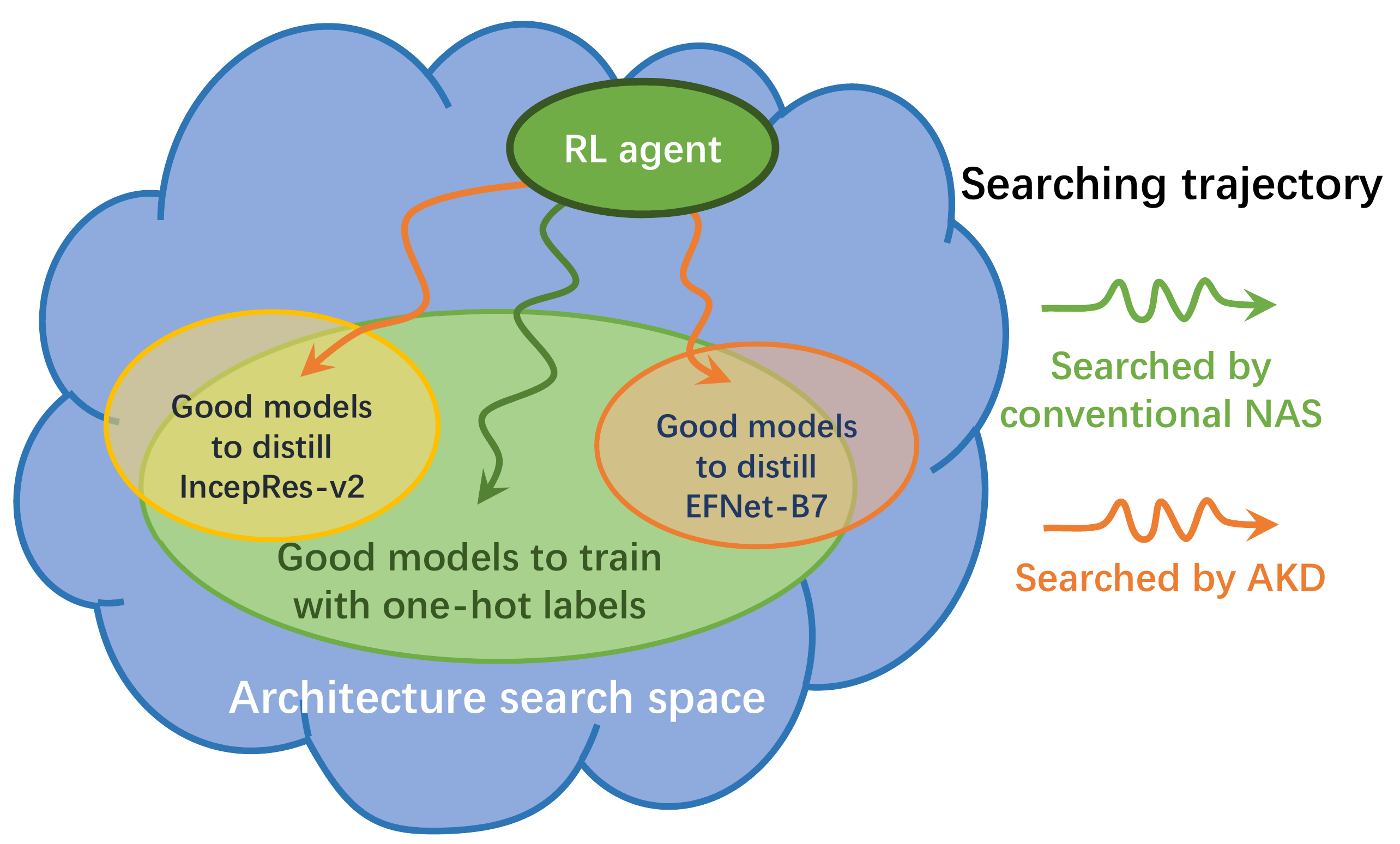}
\caption{Searching neural architectures by the proposed AKD and conventional NAS \cite{mnas19} lead to different optimal architectures.}
\label{fig:fig1}
\vspace{1pt}
\end{figure}

Aiming to reduce the need for domain knowledge and minimize human intervention, Neural Architecture Search (NAS) automates the process of neural architecture design via reinforcement learning \cite{nas_cifar17}, differentiable search \cite{liu2018darts,liu2019differentiable}, evolutionary search \cite{real2018regularized}, and other algorithms \cite{bergstra2012making}. Recently, some works have also started to explore simultaneous optimization of network parameters and architecture for a given task \cite{shin*2018differentiable}. It has been shown that NAS can discover novel architectures that surpass human-designed architectures on large-scale image classification problems \cite{real2018regularized, zoph2017learning}. 


\begin{table}
\begin{center}
\resizebox{1.0\columnwidth}{!}{
\begin{tabular}{c|ccc}
\toprule
Teachers & Student1 & Student2  & Comparison \\
\toprule
EfficientNet-B7 \cite{tan2019efficientnet} & 65.8\% & 66.6\% & student1 $<$ student2 \\
Inception-ResNet-v2 \cite{szegedy2016inceptionv4} & 67.4\% & 66.1\% & student1 $>$ student2\\
\bottomrule
\end{tabular}
}
\end{center}
	\caption{ImageNet accuracy for students with different teachers.}
\label{tab:motivate_distill}
\vskip -0.1in
\end{table}

In parallel to neural architecture advances, Knowledge Distillation (KD) \cite{hinton2015distilling}, which trains a (usually small) student neural network using the supervision of a (relatively large) teacher network, has demonstrated great effectiveness over directly training the student network using hard labels. Previous works on KD \cite{heo2019comprehensive, zagoruyko2016paying} mostly focus on transferring the teacher's knowledge to a student with a pre-defined architecture. However, we argue that the optimal student architectures for different teacher models trained on the same task and dataset might be different. To validate this argument, we study the performance of two randomly constructed student models with different teachers. As shown in Table \ref{tab:motivate_distill}, while student1 outperforms student2 with EfficientNet-B7 \cite{tan2019efficientnet} as teacher, student2 works better with the Inception-ResNet-v2 \cite{szegedy2016inceptionv4} as teacher (see  Table \ref{tab:teacher_performance} for more comprehensive studies). These results indicate that each teacher model could potentially have its own best student architecture.

Motivated by these observations, this paper proposes a new generalized approach for KD, referred to as Architecture-aware Knowledge Distillation (AKD), which finds best student architectures for distilling the given teacher model. The proposed approach (Fig \ref{fig:pipeline}) searches for student architectures using a Reinforcement Learning (RL) based NAS process \cite{nas_cifar17} with a KD-based reward function. Our results show that the best student architectures obtained by AKD achieve state-of-the-art results on the ImageNet classification task under several latency settings (Table \ref{tab:comp_sota_arch}) and consistently outperform the architectures obtained by conventional NAS \cite{mnas19} (Table \ref{tab:imagenet_latency_result}). Surprisingly, the optimal architecture obtained by AKD for the ImageNet classification task generalizes well to other tasks such as million-level face recognition (Table \ref{tab:megaface}) and ensemble learning (Table \ref{tab:mega_ensemble}). Our analysis of the neural architectures show that the proposed AKD and conventional NAS \cite{mnas19} processes select architectures from different regions of the search space (Fig \ref{fig:searching_path}). In addition, our analysis also verifies our assumption that the best student architectures for different teacher models differ (Fig \ref{fig:two_t_two_p}). To the best of our knowledge, this is the first work to investigate and utilize structural knowledge of neural architectures in KD on large-scale vision tasks.

\section{Knowledge distillation}
In this section, we present the standard KD framework and provide motivation for the proposed architecture-aware knowledge distillation approach.

\subsection{Knowledge in a neural network}
For a task with input space $\mathcal{I}$ and output space $\mathcal{O}$, an ideal model is a connotative mapping function $f:x \mapsto y,$ $x\in\mathcal{I}, y\in\mathcal{O}$ from the task's input space $\mathcal{I}$ to output space $\mathcal{O}$ that produces the correct output label for every input, and the knowledge of this model is represented by the model's conditional probability function $p(y|x)$. The knowledge of a neural network $\hat{f}:x \mapsto \hat{y}, x\in\mathcal{I}, \hat{y} \in\mathcal{O}$, trained for this task is represented by the network's conditional probability function $\hat{p}(\hat{y}|x)$. The difference between $\hat{p}(\hat{y}|x)$ and $p(y|x)$ is the dark part of the neural network's knowledge.


Researchers have tried to understand this dark part in different ways.
A recent work~\cite{phuong2019towards} showed that the data geometry, such as margin between classes, can strongly benefit KD in the case of a single-layer linear network. While the ideal one-hot output $y$ constrains the angular distances between different classes to be the same $90^{\circ}$, the soft probability output $\hat{y}$ has a learning-friendly dynamic metric: more similar classes/samples should have smaller angular distances. This latent structure of the soft label distribution contributes to the dark knowledge, and supervising the student model by the soft label distribution would work better in comparison to one-hot labels. Hinton et al.~\cite{hinton2015distilling} proposed the knowledge distillation framework, which tries to distil the soft distribution of a teacher into a student neural network with a pre-defined architecture. 

\subsection{Na\"{i}ve distillation}
Interest in KD increased following Hinton et al.~\cite{hinton2015distilling}, who demonstrated a method called \textit{dark} knowledge distillation, in which a student model trained with the objective of matching full softmax distribution of the teacher model. Commonly, the teacher is a high-capacity model with formidable performance, while the student network is compact. By transferring knowledge, one hopes to benefit from both the student's compactness and the teacher's capacity. While this strategy has been widely adopted, especially for edge devices that require low memory or fast execution, there are few systematic and theoretical studies on how and why knowledge distillation improves neural network training. \cite{hinton2015distilling} suggest that the success of KD depends on the distribution of logits of the wrong responses, that carry information on the similarity between output categories. ~\cite{furlanello2018born} argue that soft-target distribution acts as an importance sampling weight based on the teacher's confidence in its maximum value. \cite{zhang2017deep} analyzed knowledge distillation from the posterior entropy viewpoint claiming that soft-targets bring robustness by regularizing a much more informed choice of alternatives than blind entropy regularization. Nevertheless, to our knowledge, no previous works attempt to explain from the perspective of network inherent architecture, as previous efforts investigate mainly from learning theory and distillation methods.

\subsection{Teacher-Student relationship}


Although KD can be applied to any pair of teacher and student networks, a natural question is: \emph{are all student networks equally capable of receiving knowledge from different teachers?} To answer this question, we grab 8 representative off-the-shelf teacher models,  which vary in term of input size, architecture and capacity, as shown in Table~\ref{tab:teacher_student_similarity}. Subsequently, we randomly sample 5 different student architectures (which have similar performance when trained with one-hot label vectors) from the search space defined by MNAS \cite{mnas19} to investigate the role of student architecture when transferring dark knowledge.  

\begin{figure}
\centering
\includegraphics[width=1\linewidth]{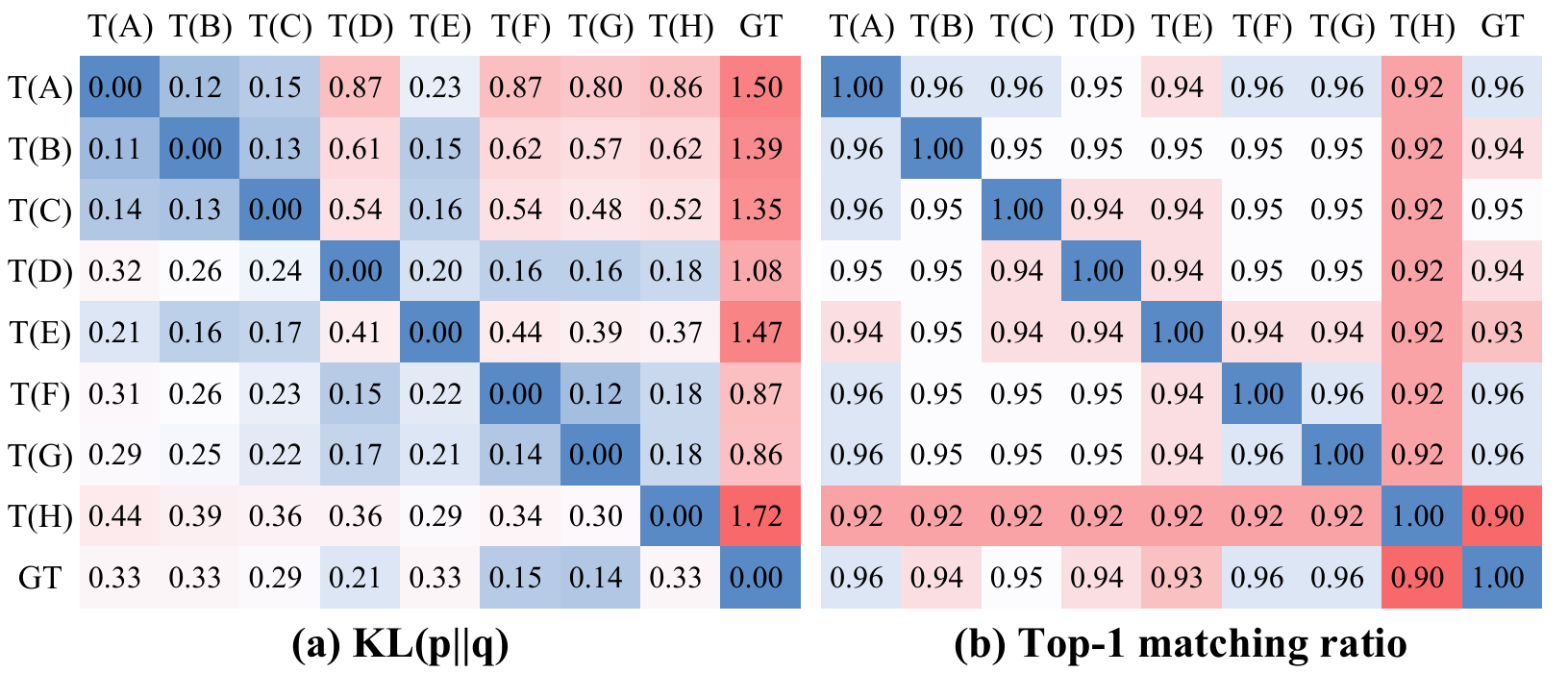}
\caption{Confusion matrix for models' outputs, p or q denotes softmax probability output, GT means one-hot label.}
\label{fig:teacher_div}
\end{figure}

Table~\ref{tab:teacher_performance} summarizes the knowledge distillation performance for each pair of teacher and student. Distilling the same teacher model to different students leads to different performance results, and no student architecture produces the best results across all teacher networks. Although the reason behind it could be complicated, we argue its not solely due to: 
\begin{itemize}
\item Distribution: Fig.~\ref{fig:teacher_div} shows T(A) \& T(B) demonstrate lowest KL divergence among many others, which means their distributions are cloest. Though distribution is only supervision from KD during learning process, in Table ~\ref{tab:teacher_performance}, $S_2$ is the pearl (best student) in the eye of T(A), whereas, $S_2$ is the last choice to T(B). Therefore it indicates that we need to disentangle distribution into finer singularity.
\item Accuracy: T(A) is considered as the most accurate model, however students fail to achieve top performance when compared with distillation by other teachers. \cite{mirzadeh2019improved} argues that the teacher complexity could hinder the learning process, as student does not have the sufficient capacity to mimic teacher's behavior, but its worth noting that $S_2$ perform significantly better than $S_1$ even though they have similar capacity. \cite{heo2019comprehensive} tends to believe the output of a high performance teacher network is not significantly different from ground truth, hence KD become less useful. However, Table ~\ref{fig:teacher_div} suggest otherwise, a lower-performance model T(F), whose distribution is closer to GT than a high-performance model T(A) did, while both of them could spot satisfied student network. 
\end{itemize}
These observations inspire us to rethink the knowledge in a teacher network, which we argue depends not only on its parameters or performance but also on its architecture. In other words, if we distill knowledge only with a pre-defined student architecture like standard KD does, it might forces the student to sacrifice its parameters to learn teacher's architecture, which end up with a non-optimal solution.

\begin{table}[t]
\begin{center}
	\footnotesize{
\begin{tabular}{ c|c|c|c}
\toprule
Tag & Model name & Input size & Top-1 accuracy \\
\toprule
T(A)  & EfficientNet-B7~\cite{tan2019efficientnet} & 600 & 84.4 \\
\midrule
T(B)  & PNASNet-large~\cite{Liu_2018_ECCV} & 331 & 82.9	74 \\
\midrule
T(C)  & SE-ResNet-154~\cite{hu2017squeezeandexcitation} & 224 & 81.33 \\
\midrule
T(D)  & PolyNet~\cite{Zhang_2017_CVPR} & 331 & 81.23 \\
\midrule
T(E)  & Inception-ResNet-v2~\cite{szegedy2016inceptionv4} & 299 & 80.217 \\
\midrule
T(F)  & ResNeXt-101~\cite{Xie_2017_CVPR} & 224 & 79.431 \\
\midrule
T(G)  & Wide-ResNet-101~\cite{zagoruyko2016wide} & 224 & 78.84 \\
\midrule
T(H)  & ResNet-152~\cite{he2015deep} & 224 & 78.31 \\
\bottomrule
\end{tabular}
}
\end{center}
	\caption{A comparison of popular off-the-shelf models, sorted by top-1 accuracy.}
\label{tab:teacher_student_similarity}
\end{table}

\begin{table}[t]
\begin{center}
	\small{
\begin{tabular}{ccccc}
\toprule
\vspace{3pt}
& \multicolumn{3}{c}{Teacher models} & \\
GT & T(A) & T(B) & T(E) & T(F) \\
\toprule
$S_3$ (65.6) & $S_2$ (66.6) & $S_3$ (66.9) & $S_1$ (67.4) & $S_5$ (67.1)\\
\midrule
$S_4$ (65.6) & $S_3$ (66.5) & $S_5$ (66.4) & $S_4$ (67.0) & $S_1$ (67.1)\\
\midrule
$S_5$ (65.5) & $S_4$ (66.3) & $S_4$ (66.1) & $S_5$ (66.9) & $S_4$ (66.6)\\
\midrule
$S_1$ (65.5) & $S_5$ (66.0) & $S_1$ (65.7) & $S_3$ (66.5) & $S_3$ (66.3)\\
\midrule
$S_2$ (65.4) & $S_1$ (65.8) & $S_2$ (65.4) & $S_2$ (66.1) & $S_2$ (66.0)\\
\bottomrule
\end{tabular}
}
\end{center}
	\caption{Distillation performance of five student architectures  from the MNAS \cite{mnas19} search space under the supervision of different teacher models. The numbers in parentheses denote the top-1 accuracy on ImageNet validation set. In each column, the student models are ordered based on their performance. Note that different teachers have different best students.}
\label{tab:teacher_performance}
\end{table}

\section{Architecture-aware knowledge distillation}
Our AKD leverages neural architecture search (NAS) to explore student architectures that can better distill the knowledge from a given teacher. In this section, we will describe our KD-guided NAS and discuss our insights.

\subsection{KD-guided NAS}

\begin{figure}
\centering
\includegraphics[width=0.85\linewidth]{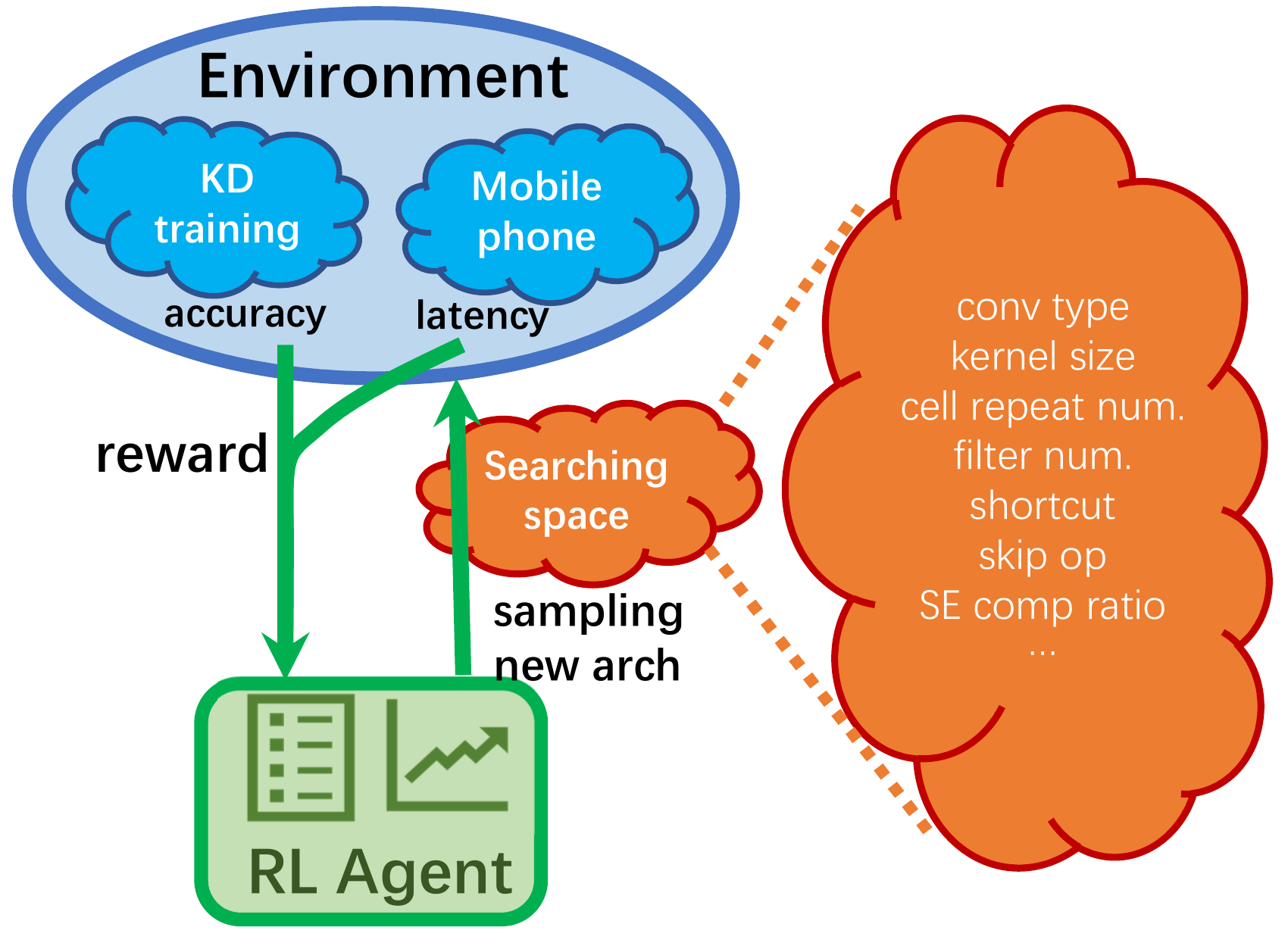}
\caption{Pipeline of searching process in AKD. there are three core components: environment, RL agent and search space. 
}
\label{fig:pipeline}
\end{figure}

Inspired by recent mobile NAS works \cite{mnas19,tan2019efficientnet}, we employ a reinforcement learning (RL) approach to search for latency-constrained Pareto optimal solutions from a large factorized hierarchical search space. However, unlike previous NAS methods, we add a teacher in the loop and use knowledge distillation to guide the search process. Fig.~\ref{fig:pipeline} shows our overall NAS flow, which consists of three major components: an RL agent for architecture search, a search space for sampling student architectures, and a training environment for obtaining KD-guided reward. 

\paragraph{RL agent: } Similar to other RL-based approaches \cite{nas_cifar17,nas_imagenet18,mnas19}, we use a RNN-based actor-critic agent to search for the best architecture from the search space. Each RNN unit determines the probability of different candidates for a search option. The RNN weights are updated using PPO algorithm \cite{ppo17} by maximizing the expected KD-guided reward.

\paragraph{Search Space: } Similar to \cite{mnas19}, our search space consists of seven predefined blocks, with each block contains a list of identical layers. Our search space allows us to search for the number of layers, the convolution and skip operation type, conv kernel size, squeeze-and-excite ratio, and input/output filter size, for each block independently.

\paragraph{KD-guided reward: } Given a sampled student architecture from our search space, we train it on a proxy task to obtain the reward. Unlike prior NAS works, we perform KD training with a teacher network to obtain KD-guided accuracy. Meanwhile, we also measure the latency on mobile devices, and use the same weighted product of accuracy and latency in \cite{mnas19} as the final KD-guided reward to approximate Pareto-optimal solutions.

\subsection{Implementation details}

Unless otherwise specified, we adopt Inception-ResNet-v2~\cite{szegedy2016inceptionv4} as the teacher model in ImageNet experiments. Our NAS process mostly follows the same settings in MNAS~\cite{mnas19}. We reserve 50K images from training set as mini-val, and use the rest as mini-train. We treat each sampled model as a student, and distill teacher's knowledge by training on the mini-train for 5 epochs, including the first epoch with warm-up learning rate, and then evaluating on the mimi-val to obtain its accuracy. We have also tried to train each sampled model for a longer time (15 epochs) and observe minor performance improvement. After that, the sampled model will be evaluated on the single-thread big CPU core of Pixel 1 phones to measure its latency. Finally, the reward is calculated based on the combination of accuracy and latency. We do not use a shared weight among different sampled models. 

We use actor-critic based RL agent for neural architecture search. In each searching experiment, the RL agent samples $\sim$10K  models. Then we pick the top models that meet the latency constrain, and train them for further 400 epochs by either distilling the same teacher model or using ground truth labels. Following common practices in KD, temperature is set to 1 and the distilling weight $\alpha$ is set to 0.9. We launch all the searching experiments on a large TPU Donut cluster~\cite{jouppi2017indatacenter}. Since each sampled model distills the teacher in an online manner, the searching time of AKD is 3$\sim$4 times longer than the conventional NAS. Each experiment takes about 5 days on 200 TPUv2 Donut devices.

\begin{figure*}
\centering
\includegraphics[width=\linewidth]{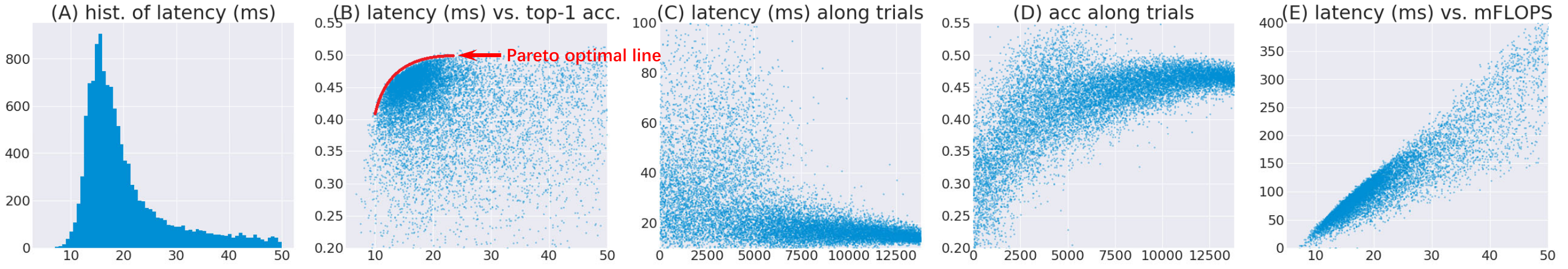}
\caption{Attributes of searching process. Each dot indicates one architecture.}
\label{fig:search_info}
\end{figure*}
\begin{figure*}
\centering
\includegraphics[width=\linewidth]{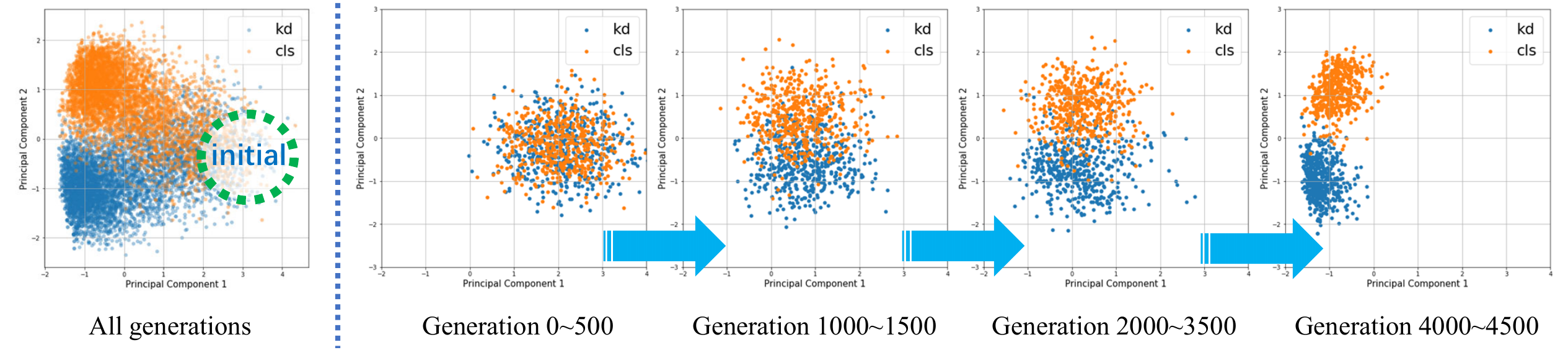}
\caption{The architecture evolves during searching. Each dot represents an architecture. Different colors indicate different NAS policies -- orange for conventional NAS and blue for AKD based NAS. PCA is used for visualization. Best view in color.}
\label{fig:searching_path}
\end{figure*}
\vspace{-5mm}
\subsection{Understanding the searching process}
For the better readability, we denote the family of architecture searched by KD-guided reward as \textbf{AKDNet}, and the family searched by classification-guided reward as \textbf{NASNet} in the following paragraphs. Note that in each comparable experiment we initialize the RL agent with the same random weights and latency target. 

Fig.~\ref{fig:search_info} presents some interesting statistical results of an AKD searching process, where the latency target is set to 15ms. We can see that with the soft reward design by ~\cite{mnas19}, all generations of sampled models have the most dense distribution around the target latency in Fig.~\ref{fig:search_info} (A). Fig.~\ref{fig:search_info} (B) illustrates the latency-accuracy trade-off among all sampled models, where the Pareto optimal curve is highlighted. All the quantitative results in the latter sections are produced by the architectures close to this curve. Fig.~\ref{fig:search_info} (C) and (D) show how the latency and accuracy of sampled models changes along the searching generations. It is interesting to see the RL agent firstly targets on higher accuracy, and does not start finding efficient architectures until $\sim$5K generations. Fig.~\ref{fig:search_info} (E) shows the correlation between latency and FLOPS. Intriguingly, consider a same latency (in vertical), we can see AKD favor models with higher FLOPS, which is quite reasonable. More analysis for FLOPS will be shown in Sec.~\ref{sec:latency_vs_flops}.

To better investigate how different the AKDNet and NASNet are, Fig.~\ref{fig:searching_path} shows how the sampled architecture evolves in the search space. In the first frame, the `initial' circle denote the architectures sampled by a randomly initialized RL agent. The latter 4 frames show 4 groups of 500 models sampled at different stages. The groups searched by AKD and conventional NAS are separable, indicating the distillation task introduces some new \textbf{undiscovered information} to rectify the student architecture evolves into another direction. We demonstrate it \textit{\textbf{is}} the structural knowledge of the teacher model in the next section.

\section{Understanding the structural knowledge}
\label{sec:understand_structural_knowledge}
\subsection{Existence of structural knowledge}

The main assumption of AKD is the knowledge behind the teacher network includes structural knowledge which sometimes can hardly be transferred to parameters. Although the universal approximation theorem~\cite{hornik1991approximation} asserts that a simple two-layer neural network can approximate any continuous function in an ideal condition, \cite{hornik1991approximation} assumes the size of the universal model can be extremely large, which is unpractical. 

In this section we investigate the existence of structural knowledge by answering two questions: 
\begin{enumerate}
\vspace{-2mm}
  \item If two identical RL agents perform AKD on two different teacher architectures, will they converge to different area in search space?
 \vspace{-2mm}
  \item If two different RL agents perform AKD on the same teacher, will they converge to similar area?
\end{enumerate}

\vspace{-5mm}
\paragraph{Different teachers with same RL agent}

\begin{figure}
\centering
\includegraphics[width=.9\linewidth]{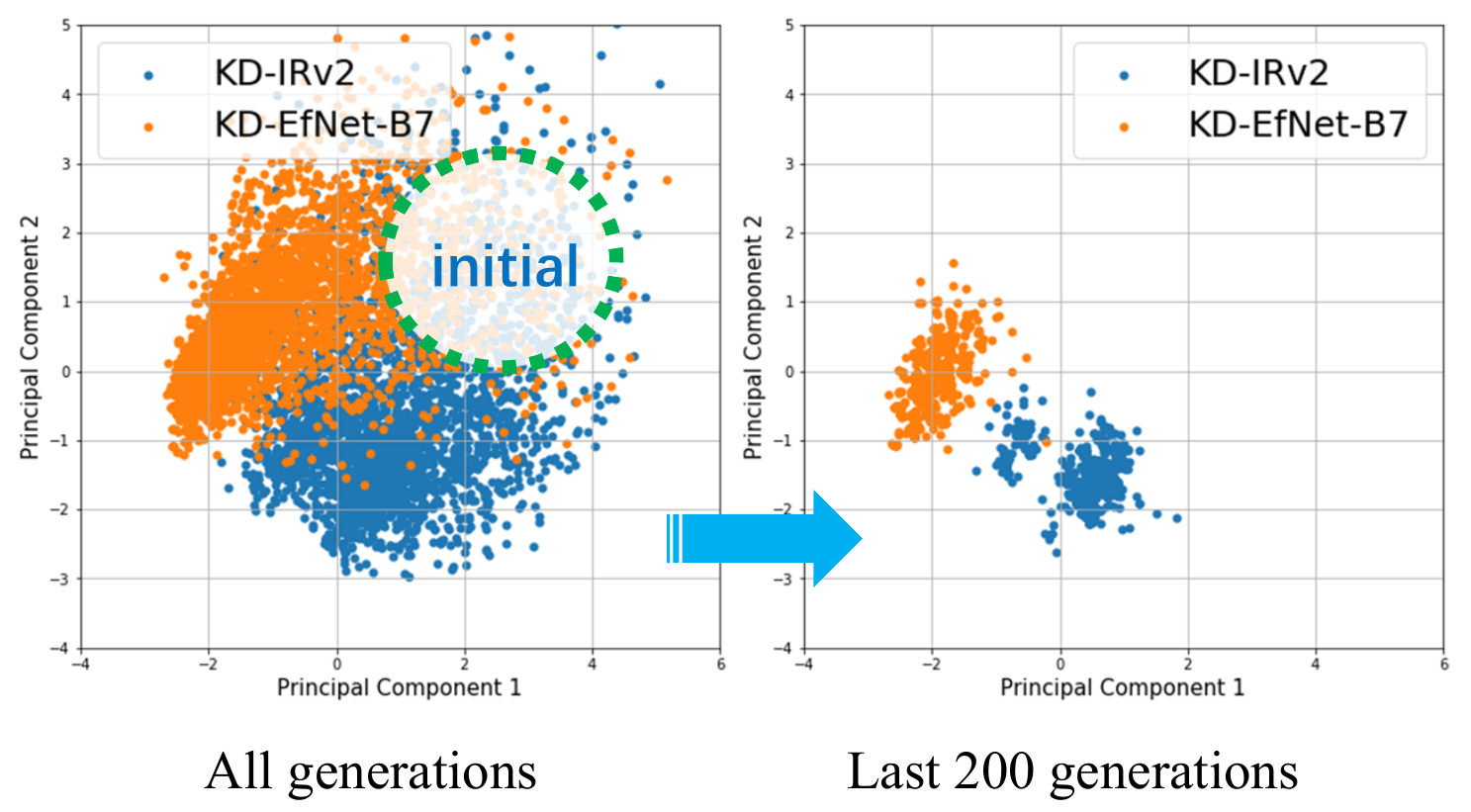}
\caption{All generations searched by AKD on two different teachers. Their final generations converge into different areas. This proves the structural knowledge does exist in the teacher model.}
\label{fig:two_t_two_p}
\end{figure}

We answer the first question by launching two AKD searching processes. The teacher models are the off-the-shelf Inception-ResNet-v2~\cite{szegedy2016inceptionv4} and EfficientNet-B7~\cite{tan2019efficientnet}, respectively. Besides the teacher model, all the other settings, random seed, latency target and mini-val data are fixed to the same setting. Fig.~\ref{fig:two_t_two_p} shows the initial distribution (in the green dot circle), all generations (left) and the final converged architectures (right). The final optimal architectures are clearly separable.

\vspace{-2mm}
\paragraph{Same teacher with different RL agent}
\begin{figure}
\centering
\includegraphics[width=.9\linewidth]{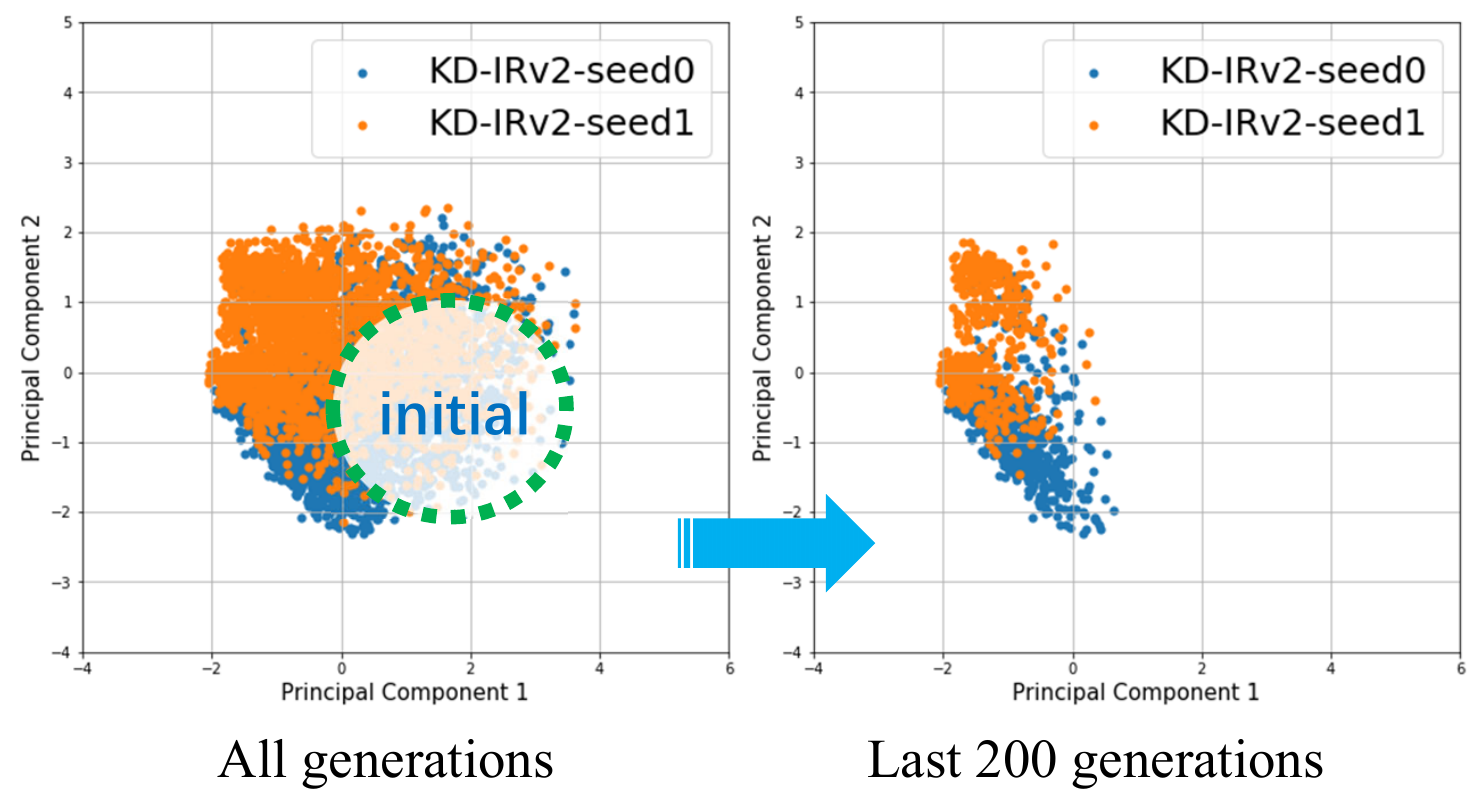}
\caption{All generations searched by AKD on the same teacher model but  different RL agents. Their final generations converge into the same area. Note that there are a large amount of blue dots overlapped by the orange dots. Best view in color.}
\label{fig:one_t_two_p}
\end{figure}

We further doubt whether the separability is produced by the random factor lies in the RL searching program. To answer this question, we launch two AKD searching programs for the same teacher model. In the two experiments we set different random seeds for the RL agent and different splits of mini-train / mini-val. These changes will introduce significant random factors for the initial population and reward distribution. We denote them as \texttt{KD-IRv2-seed0} and \texttt{KD-IRv2-seed1} in Fig.~\ref{fig:one_t_two_p}. Not surprisingly they finally converge to the close area.

These two experiments jointly prove the structural knowledge does exist in a neural network.

\paragraph{Difference between AKDNet and NASNet}
\begin{figure*}
\centering
\includegraphics[width=\linewidth]{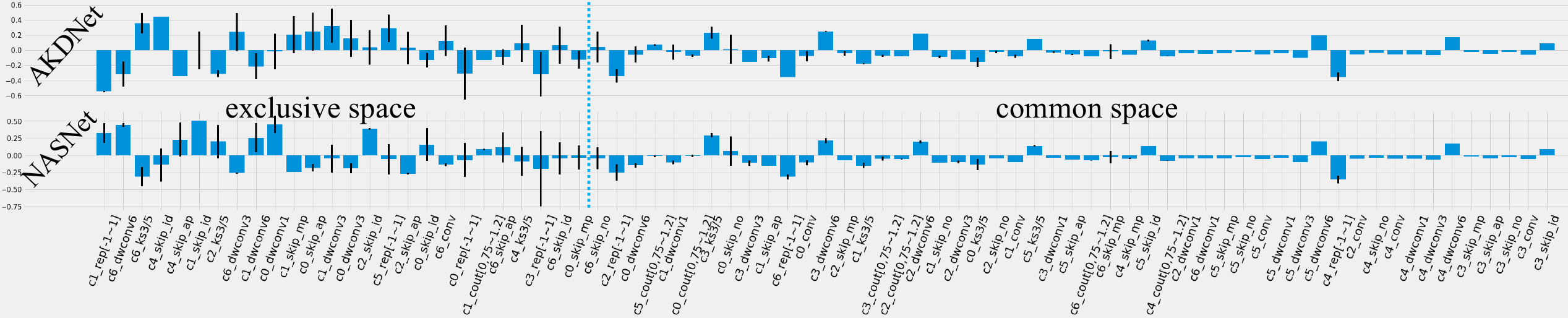}
\caption{Mean and standard deviation of probability of each operator. Calculated by the 100 optimal AKDNet and NASNet and sorted by the significance of difference.}
\label{fig:arch_diff}
\end{figure*}
In the previous paragraph we investigated the structural knowledge. But it is still too abstract to make a visual comprehension. In this work we try to get some insight by comparing the statistical divergence between the architecture family of AKDNet and NASNet. We first expand the search space to make it continuous: \textit{e.g.} the original `skip\_op' dimension has 4 values for 4 denoting skip operator, and we need to expand it to 4 one-hot dimensions. In this way the original 35-dim space is expanded to 77-dim. After that, we calculate the probability of each operator among the top 100 optimal architectures of AKDNet and NASNet respectively, constrained by a range of latency, and get the difference between them (AKDNet minus NASNet). Result is shown in Fig.~\ref{fig:arch_diff}. Some interesting things are: compared with NASNet, students of Inception-ResNet-v2 favor the larger kernel size but small expand ratio of \texttt{depth-wise conv}, and tend to reduce the layer number at the beginning of the network.

\subsection{AKDNet becomes KD-friendly along searching}

\begin{table}[t]
\begin{center}
	\small{
\begin{tabular}{ c|cccc}
\toprule
\vspace{3pt}
Generation & initial & $\sim$1k & $\sim$3k & $\sim$10k \\
\midrule
Winning ratio & 10 / (20-2) & 12 / 20 & 16 / 20 & \textbf{18 / 20} \\
\midrule
Average gain & -0.07 & 0.10 & 0.46 & \textbf{1.05} \\
\bottomrule
\end{tabular}
}
\vspace{10pt}
\caption{Relative performance gain between KD(AKDNet) and KD(NASNet) during different searching stages.}
\label{tab:rg}
\end{center}
\end{table}

A good metric to measure how AKDNet improves KD result is the relative performance gain. It is defined as:
\begin{equation}
\scriptsize{\rm
    [KD(AKDNet)-CLS(AKDNet)]-[KD(NASNet)-CLS(NASNet)],
}
\end{equation}
in which `KD' and `CLS' denote the final performance of the network which will be trained for 400 epochs with KD or classification loss. 

We randomly sample 80 architectures from 4 stages among 10,000 generations for both NAS and AKD, and calculate the relative gain in each stage. Tab.~\ref{tab:rg} shows the results. The wining ratio indicates how many AKDNet models beat the NASNet. It can be observed that the average gain and the wining ratio go up along the searching process. AKDNet becomes more and more friendly to the teacher model distillation.

\section{Preliminary experiments on ImageNet}


\begin{table}[t]

\begin{center}
	\footnotesize
\begin{tabular}{ c|cccc}
\toprule
\vspace{3pt}
Latency & searching by & training by & top-1 & top-5\\
\toprule
%

\rowcolor{mygray}\cellcolor{white}& hard label & hard label & 59.73 & 81.39 \\
\cmidrule{2-5}
							& hard label & distillation & 63.9 & 84.26  \\
\cmidrule{2-5}
\rowcolor{mygray}\cellcolor{white}& distillation & hard label & 61.4 & 83.1 \\
\cmidrule{2-5}
\multirow{-5}{*}{\cellcolor{white}15$\pm$1 ms}& \textbf{distillation} & \textbf{distillation} & \textbf{66.5} & \textbf{87.5} \\

\midrule


\rowcolor{mygray}\cellcolor{white}& hard label & hard label & 67.0 & 87.4 \\
\cmidrule{2-5}
							& hard label & distillation & 68.1 & 88.0  \\
\cmidrule{2-5}
\rowcolor{mygray}\cellcolor{white}& distillation & hard label & 67.2 & 87.5 \\
\cmidrule{2-5}
\multirow{-5}{*}{\cellcolor{white}25$\pm$1 ms}& \textbf{distillation} & \textbf{distillation} & \textbf{69.6} & \textbf{89.1} \\
\midrule

\rowcolor{mygray}\cellcolor{white}& hard label & hard label & 73.0 & 92.1 \\
\cmidrule{2-5}
							& hard label & distillation & 74.7 & 92.54  \\
\cmidrule{2-5}
\rowcolor{mygray}\cellcolor{white}& distillation & hard label & 73.6 & 92.2 \\
\cmidrule{2-5}
\multirow{-5}{*}{\cellcolor{white}75$\pm$1 ms}& \textbf{distillation} & \textbf{distillation} & \textbf{75.5} & \textbf{93.1} \\
\bottomrule
\end{tabular}
\end{center}
\caption{Ablation study on ImageNet.}
\label{tab:imagenet_latency_result}
\end{table}
Table~\ref{tab:imagenet_latency_result} shows the performance of our models on ImageNet \cite{krizhevsky2012imagenet}, with several target latencies. After search, we pick the top-performing AKDNet and NASNet, and train them until converge by distillation or hard label respectively. As shown in the table, AKDNets consistently outperform NASNet in all settings by a large margin, e.g, AKDNet achieves 61.4\% top-1 / 83.1\% top-5 accuracy compared to NASNet's 59.73\% top-1 / 81.39\% top-5 accuracy with 15ms latency. Moreover, If we further train them with distillation, AKDNet improve \(61.4\%\rightarrow 66.5\%\) while NASNet \(59.73\%\rightarrow63.9\%\), which suggests that our framework capable to find architecture that good for distillation. 

\subsection{Transfer AKDNet to advanced KD methods}

\begin{table}[t]
\begin{center}
\resizebox{1.0\columnwidth}{!}{
\begin{tabular}{ c|c|ccc}
\toprule
Training & Latency & \multicolumn{3}{c}{Advanced KD method}\\
method&  & TA-KD & RCO-KD & CC-KD \\
\toprule
MNet-v2 w/o KD &  & \multicolumn{3}{c}{65.4} \\
\cmidrule{3-5}
MNet-v2 w/ KD & \multirow{-2}{*}{33ms} & 67.6 $\uparrow$\textbf{2.2} & 68.2 $\uparrow$\textbf{2.8} & 67.7 $\uparrow$\textbf{2.3} \\
\midrule
AKDNet-M w/o KD &  & \multicolumn{3}{c}{68.9} \\
\cmidrule{3-5}
AKDNet-M w/ KD & \multirow{-2}{*}{32.8ms} & 72.0 $\uparrow$\textbf{3.1} & 72.4 $\uparrow$\textbf{3.5} & 72.2 $\uparrow$\textbf{3.3} \\
\bottomrule
\end{tabular}
}
\end{center}
	\caption{AKDNet transfers to other advanced KD method. `MNet-v2' denotes the MobileNet-v2 0.5$\times$. The 'M' in AKDNet-M denotes the 32.8ms version of ADKNet. Even with a much higher baseline, AKDNet consistently brings considerable gain ($\sim1\%$) under each KD method compared to MobileNet-v2.}
\label{tab:akdnet_with_kd}
\end{table}

Since AKD uses the same training method during searching and final training, it is interesting to investigate whether AKD overfits the original KD policy, such as the algorithm and hyper-parameters. So we verify it by transferring the Pareto optimal architecture at $\sim$33 ms (denoted as AKDNet-M) to the three recent state-of-the-art KD methods, TA-KD~\cite{mirzadeh2019improved}, RCO-KD~\cite{jin2019knowledge} and CC-KD~\cite{peng2019correlation}. We select the 33 ms to match the most common settings in these three works. Note that since the TA-KD does not have experiments on ImageNet and MobileNet-v2, we use a new implementation that has 3 TA~\cite{mirzadeh2019improved} networks which are channel-expanded version of the target student architecture. 

Results are shown in Tab.~\ref{tab:akdnet_with_kd}. It can be observed that even with a quite strong baseline when training without KD, AKDNet-M still can get more improvement under all KD methods.

\subsection{Compare with SOTA architectures}

In the previous experiments, we focus on the architectures sampled or searched in the same search space. This may introduces bias into our conclusion. Now we evaluate the performance gain produced by distillation under different architecture settings. In each setting, we first select a state-of-the-art architecture, either designed by human or searched by NAS. Then we select an architecture in the Pareto optimal set of AKD that has the most similar latency with that of the SOTA model. We compare how much improvement can be made when using them to distill the same teacher model (Inception-ResNet-v2). MobileNet-v2~\cite{mobilenetv218}, MNasNet~\cite{mnas19} and MobileNet-v3~\cite{mobilenetv319} are compared in Tab.~\ref{tab:comp_sota_arch}. We adopt the original KD and the RCO-KD to show the consistent conclusion.

It is worth to mention that this work does not aim at beating the other backbone designing for classification, but finding an architecture that can get more gain when training by KD. However not surprisingly the Tab.~\ref{tab:comp_sota_arch} also proves the AKDNet can achieve great performance when trained without KD thanks to the good search space.

\begin{table}[t]
\footnotesize
\begin{center}
\resizebox{1.0\columnwidth}{!}{
\begin{tabular}{ c|cccc}
\toprule
\vspace{3pt}
Latency & architecture  & with KD? & top-1 & top-5 \\
\toprule

\rowcolor{mygray}\cellcolor{white}& AKDNet &  & 61.4 & 83.1\\
							& AKDNet &  \checkmark & $\uparrow$2.6 & $\uparrow$3.24 \\
							& AKDNet &  RCO-KD & $\uparrow$3.1 & $\uparrow$3.8 \\
\cmidrule{2-5}
\rowcolor{mygray}\cellcolor{white}& MNet-v2-a &  & 59.2 & 79.8  \\
                            & MNet-v2-a & \checkmark & $\uparrow$1.4 & $\uparrow$2.1  \\
\cmidrule{2-5}
\rowcolor{mygray}\cellcolor{white}& MNASNet-a &  & 62.2 & 83.5 \\
                            & MNASNet-a & \checkmark & $\uparrow$1.49 & $\uparrow$2.3  \\
\cmidrule{2-5}
\rowcolor{mygray}\cellcolor{white}& MNet-v3-a &  & 64.1 & 85.0 \\
\multirow{-10}{*}{\cellcolor{white}15$\sim$20 ms}& MNet-v3-a & \checkmark & $\uparrow$1.3 & $\uparrow$ 2.2  \\

\midrule

\rowcolor{mygray}\cellcolor{white}& AKDNet &  & 67.2 & 87.5\\
							& AKDNet &  \checkmark & $\uparrow$2.4 & $\uparrow$1.6 \\
							& AKDNet &  RCO-KD & $\uparrow$2.8 & $\uparrow$1.5 \\
\cmidrule{2-5}
\rowcolor{mygray}\cellcolor{white}& MNASNet-b &  & 66.0 & 86.1 \\
\multirow{-6}{*}{\cellcolor{white}25$\sim$27 ms}& MNASNet-b & \checkmark & $\uparrow$1.1 & $\uparrow$0.6  \\

%
%

\bottomrule
\end{tabular}
}
\end{center}
	\caption{Comparison to the state-of-the-art architecture. AKDNet has the comparable results when trained without KD, and its performance get improved by a large margin when trained with KD. We reduce the channels of all the compared model with the same reduction ratio in each layer to obtain a comparable latency. We select the AKDNet to match the lower bound latency in each comparable group.}
\label{tab:comp_sota_arch}
\end{table}

\subsection{Latency vs. FLOPS}
\label{sec:latency_vs_flops}

As some of recent works~\cite{mnas19} argue that latency can be a better metric to describe the computational cost compared with other proxy metrics like FLOPS, we constrain the NAS process by a soft target of latency in all settings. However it is still interesting to see the exact correlation between latency and FLOPS in our search space. 
Figure~\ref{fig:search_info} (e) presents how FLOPS changes \textit{w.r.t.} the latency, which covers around 14,000 models in the searching trajectory for the 15ms-target. By performing a simple linear regression, we observe that the FLOPS and latency are empirical linearly correlated as follows:
\vspace{-2mm}
$${\rm 3.4 \times (latency-7)\leq mFLOPS \leq 10.47 \times (latency-7)}$$
\vspace{-7mm}

The variance becomes larger when the model goes up. To verify whether the conclusion still holds on searching by a slack FLOPS constraint, we replace the latency term in our reward function by a FLOPS term, and set the target to 300 mFLOPS. Tab.~\ref{tab:flops} shows AKD leads to consistent improvement regardless of how we define the searching target and how we distill the teacher.

\begin{table}[t]

\begin{center}
	\small{
\begin{tabular}{ c|cccc}
\toprule
\vspace{3pt}
 & searching by  & training by & top-1 & top-5 \\
\toprule

\cellcolor{white}& hard label & hard label & 69.92 & 89.1 \\
\cmidrule{2-5}
\multirow{-2}{*}{NASNet }							& hard label & distillation & 71.2 & 90.4  \\
\midrule
 & distillation & hard label & 70.0 & 89.4 \\
\cmidrule{2-5}
                            & \textbf{distillation} & \textbf{distillation} & \textbf{72.1} & \textbf{91.7} \\
\cmidrule{2-5}
\multirow{-3}{*}{AKD}& \textbf{distillation} & \textbf{RCO-KD} & \textbf{73.0} & \textbf{92.2} \\

\bottomrule
\end{tabular}
}
\vspace{5pt}
\caption{ImageNet Performance with FLOPS-constrained AKD.  All settings are the same as in Tab.~\ref{tab:imagenet_latency_result} except using FLOPS (rather than latency) as the reward. Target FLOPS is about 300M. Despite using different reward, AKD consistently improves performance over NASNet.
}
\label{tab:flops}
\end{center}
\end{table}

\section{Towards million level face retrieval}

\begin{table}[t]
\begin{center}
	\small{
\begin{tabular}{ c|c|cccc}
\toprule
\vspace{3pt}
Training & KD & \multicolumn{4}{c}{Distractor num.} \\
method & method & 1e3 & 1e4 & 1e5 & 1e6 \\
\toprule
Teacher  & - & 99.56 & 99.3 & 99.0 & 98.2\\
\midrule
MNet-v2  & - & 91.49 & 84.45 & 75.6 & 65.9 \\
MNet-v2  & CC-KD  & 97.93 & 95.76 & 91.99 & 86.29 \\
MNet-v2  & RCO-KD & 98.29 & 95.01 & 90.97 & 85.9 \\
\midrule
\bf AKDNet-M  & - & 93.8 & 86.4 & 78.2 & 68.6 \\
\bf AKDNet-M  & CC-KD & 98.26 & 97.48 & 93.85 & 88.41 \\
\bf AKDNet-M  & RCO-KD & 98.42 & 97.56 & 94.1 & 90.2 \\
\bottomrule
\end{tabular}
}
\end{center}
	\caption{Transfer the AKDNet on MegaFace. The teacher model in all KD settings is Inception-ResNet-v2. }
\label{tab:megaface}
\end{table}

It is much harder to learn a complex data distribution for a tiny neural network, but some previous works~\cite{jin2019knowledge,peng2019correlation} have shown that a huge improvement can be achieved by introducing KD in  complex metric learning. We adopt the most challenging face retrieval benchmark MegaFace~\cite{kemelmacher2016megaface} to verify the transferability of our searched models. The MegaFace contains 3,530 probe faces and more than 1 million distractor faces. The target is to learn a neural network on an irrelevant training set like MS-Celeb-1M~\cite{guo2016ms} and test the model on MegaFace in a zero-shot retrieval manner.

We first train an Inception-ResNet-v2 on the MS-Celeb-1M~\cite{guo2016ms}, pre-processed by RSA~\cite{liu2017recurrent}, and directly distill its representation to the AKDNet-M (described in Table \ref{tab:akdnet_with_kd}). The distillation setting between ImageNet and face retrieval has a little bit difference. Since MegaFace is an open-set zero-shot benchmark, we just distill the output feature of the penultimate layer by minimizing the mean square error.
It is worth emphasizing that the AKDNet-M is searched on the ImageNet with Inception-ResNet-v2 as a teacher model. We hope it can benefit from the learned structural knowledge and directly transfer it to the face retrieval task. Respecting the common setting in this area, we report the Top-1 accuracy under different number of distractors when performing retrieval. Results are shown in Tab.~\ref{tab:megaface}. With a much higher baseline performance, AKDNet further achieves significant improvement on two different KD methods.

\section{Neural architecture ensemble by AKD}
The original knowledge distillation~\cite{hinton2015distilling}  can significantly improve the single model performance by distilling the knowledge in an ensemble of models. It is skeptical whether the AKDNet over-fits the teacher architecture and whether it still works well when the teacher model is an ensemble of multiple models whose architectures are different. In Tab.~\ref{tab:mega_ensemble}, we show 5 current state-of-the-art models on MegaFace and use their ensemble to be the teacher. We can observe that AKDNet can still leverage other teachers' knowledge, and obtain  consistent improvements with either original-KD or RCO-KD compared to the hard label training without teacher.

\begin{table}[t]
\begin{center}
	\footnotesize{
\begin{tabular}{ c|c|c|ccc}
\toprule
\vspace{3pt}
Training &  & Training  & \multicolumn{2}{c}{Distractor num.} \\
Method & \multirow{-2}{*}{Architecture} & method   & 1e5 & 1e6 \\
\toprule
\multirow{5}{*}{Ensemble Teacher} & HRNet-w48~\cite{WangSCJDZLMTWLX19} & \multirow{5}{*}{-} & \multirow{5}{*}{99.6} & \multirow{5}{*}{99.3} \\
  & + R100~\cite{deng2019arcface} & & & \\
  & + EPolyFace~\cite{liu2019towards} & & & \\
  & + IncRes-v2~\cite{szegedy2017inception} & & & \\
  & + SE154~\cite{hu2018squeeze} & & & \\
\midrule
AKDNet-M  & AKDNet &  hard label  & 78.2 & 68.6 \\
AKDNet-M  & AKDNet & original-KD & 94.9 & 90.1 \\
AKDNet-M  & AKDNet & RCO-KD & \textbf{95.7} & \textbf{90.9} \\
\bottomrule
\end{tabular}
}
\end{center}
	\caption{AKD with ensemble teacher. We ensemble 5 state-of-the-art models, and use it as a teacher to compare hard label, original-KD, and RCO-KD. Our AKDNet-M achieves consistent improvements when teacher is used (either in original-KD or RCO-KD).}
\label{tab:mega_ensemble}
\end{table}

\section{Conclusion \& further thought}

This paper is the first that points out the significance of structural knowledge of a model in KD, motivated by the inconsistent distillation performance between different student and teacher models. We raise the presumption of structural knowledge and propose a novel RL based architecture-aware knowledge distillation method to distill the structural knowledge into students' architecture, which leads to surprising results on multiple tasks. Further, we design some novel approach to investigate the NAS process and experimentally demonstrate the existence of structural knowledge.

The optimal student models in AKD can be deemed as the most structure-similar to the teacher model. This implies a similarity metric of the neural network may occur but never be discovered. It is interesting to see whether we can find a new metric space that can measure the similarity between two arbitrary architectures. If so, it would nourish most of the areas in machine learning and computer vision.

{\small
\bibliographystyle{ieee_fullname}
\bibliography{egbib}
}

\end{document}